\documentclass[10pt,twocolumn,letterpaper]{article}
\usepackage{CJKutf8}
\pdfoutput=1
\usepackage{titlesec}
\usepackage{stfloats}
\usepackage[cvprfinal]{cvpr}
\usepackage{steinmetz}
\usepackage{graphicx}
\usepackage[numbers,sort&compress]{natbib}
\usepackage{subcaption}
\usepackage{amsmath}
\usepackage{amssymb}
\usepackage{booktabs}
\usepackage{cite}
\usepackage{indentfirst}
\usepackage{float}
\usepackage{stringstrings}
\AtBeginDocument{\begin{CJK}{UTF8}{gbsn}}
	\AtEndDocument{\end{CJK}}
\usepackage[pagebackref,breaklinks,colorlinks]{hyperref}

\usepackage[capitalize]{cleveref}
\crefname{section}{Sec.}{Secs.}
\Crefname{section}{Section}{Sections}
\Crefname{table}{Table}{Tables}
\crefname{table}{Tab.}{Tabs.}

\usepackage{sectsty}
\sectionfont{\centering}

\def\cvprPaperID{*****} % *** Enter the CVPR Paper ID here

\def\confName{Arvix}
\def\confYear{2023}

\begin{document}
\newcommand\degree{^\circ}

\title{\LARGE Modular Anti-noise Deep Learning Network for Robotic Grasp Detection Based on RGB Images}  % **** Enter the paper title here

\author{Zhaocong Li\\
Guangdong University of Technology\\
{\tt\small 3120004974@mail2.gdut.edu.cn}
}

\maketitle
\begin{abstract}
   Abstract— \emph{While traditional methods relies on depth sensors, the current trend leans towards utilizing cost-effective RGB images, despite their absence of depth cues. This paper introduces an interesting approach to detect grasping pose from a single RGB image. To this end,  we propose a modular learning network augmented with grasp detection and semantic segmentation, tailored for robots equipped with parallel-plate grippers. Our network not only identifies graspable objects but also fuses prior grasp analyses with semantic segmentation, thereby boosting grasp detection precision. Significantly, our design exhibits resilience, adeptly handling blurred and noisy visuals. Key contributions encompass a trainable network for grasp detection from RGB images, a modular design facilitating feasible grasp implementation, and an architecture robust against common image distortions. We demonstrate the feasibility and accuracy of our proposed approach through practical experiments and evaluations. }
\end{abstract}
%%%%%%%%% BODY TEXT - ENTER YOUR RESPONSE BELOW
\section{\Large I\large{NTRODUCTION}}

In the rapidly advancing realm of robotics, the automation of object grasping emerges as a significant frontier. Robots, especially those equipped with mechanical arms, are increasingly expected to interact seamlessly with their environment, necessitating the ability to efficiently grasp everyday objects. However, achieving this in cluttered spaces presents a challenge, one that revolves around designing an effective processor architecture attuned to such demands. Traditional methods have often leaned on depth information from specialized sensors to achieve this. Still, the present research landscape is shifting towards a reliance on standard RGB images—readily available and cost-effective visual data devoid of depth information.

The use of RGB images, while economical and widespread, introduces unique complications. Absent the depth cues, these images demand intricate algorithms to interpret the visual intricacies, especially when robots are presented with images marred by noise or blurriness due to transmission glitches or environmental factors. These very challenges underline the achievements of recent studies in grasp detection. For instance, Yang et al. \cite{TaskOrient} employed a CRF-based semantic model in their network, significantly enhancing grasp detection in scenarios where objects are frequently obscured. Other notable contributions, like \cite{MultiTask} and \cite{SingleMulti}, have integrated object detection with grasp detection, refining the robot's ability to adjust its grasp in multifaceted environments. The role of semantic segmentation, as evidenced by works such as Ainetter et al. \cite{E2E_GD_SS}, proves instrumental, simplifying the process of identifying distinct objects and associating grasp attempts correctly. Yet, the inherent challenges persist. Occlusions in multi-object scenes often mask crucial object data, and real-world conditions further exacerbate these issues, presenting distorted, noisy, or blurry images to robots.

Addressing these challenges, this paper unveils an innovative solution: an end-to-end deep neural network fortified with semantic segmentation, specifically crafted for robots with parallel-plate grippers. Beyond its primary capability to identify graspable objects, our network integrates a refinement module, harmonizing results from earlier grasp determinations with semantic segmentation, leading to a discernible enhancement in grasp detection accuracy. Recognizing the pervasive issue of image imperfections, we've also optimized our network, enabling it to adeptly manage blurred inputs and exhibit a commendable robustness against noise-laden visuals.

The main contributions of our paper are as followed:
\begin{enumerate}
   \item A trainable neural network that can be used for graspable object detection from a single RGB image.
   \item A modular network design which integrates multiple training components, yielding promising feasible grasp detection implementation. 
   \item A robust architecture which is able to process blurry images with Gaussian or salt-and-pepper noise.
\end{enumerate}

The proposed network accepts RGB images as input and generates refined grasp candidates in the end. It is revealed by Lenz et al. \cite{Lenz_grasp} that a five-dimensional rectangular representation could be used for robotic grasp detection. Thus, we apply this promising representation for the candidates in this paper. A grasp candidate \textbf{g} is defined as 
\begin{equation}
   \mathbf{g}=(x,y,\omega,h,\theta),
   \label{equ_g_define}
\end{equation}  
where $x$ and $y$ represent the location of the center of the grasp candidate, $\omega$ and $h$ describe the width and height, and $\theta$ is the orientation of the rotated bounding box. 

%-------------------------------------------------------------------------% 
\section{\Large R\large{ELATED}\Large{ W}\large{ORK}}
\textbf{Grasp detection }has been developing for over two decades. Broadly speaking, the survey papers \cite{bicchi2000robotic, survey3_3D, du2021vision} provide a thorough context for this widely studied field: Bicchi et al. \cite{bicchi2000robotic} conduct a comprehensive survey of the research conducted in robotic grasping from approximately 1980 to 2000, with a specific focus on progress made in a theoretical framework and analytical outcomes within this domain; in a similar vein, the work by Sahbani et al. \cite{survey3_3D} offers an overview of algorithms that generate autonomous gripper-based object grasps in 3D views; furthermore, \cite{du2021vision} provides a comprehensive review of vision-based robotic grasping, categorizing it into three fundamental tasks: object pose estimation, object localization, and grasp estimation. Early work on visual-based system combining grippers with a camera goes back to \cite{kamon1996learning} which stores location parameters as well as quality parameters for each grasp. It reveals that learning-based methods could generate promising outcome for robotic grasping as well. Therefore, machine learning has been also developing alongside grasping manipulations. \cite{saxena2008robotic} utilizes Guassian density function, logistic regression and YCbCr color space to predict grasp candidates, using the appropriate grasping point. As machine learning develops, deep learning also makes breakthrough with the advent of CNN and the high performance GPUs which support the large-scale and complex training missions for networks. \cite{ghazaei2015exploratory} employed a CNN architecture to classify the objects in the COIL100 database to evaluate if an intelligent vision system can be used to enhance the grip functionality of prosthetic hands. \cite{pinto2016supersizing} enlarged the size of a dataset to the 40 times more than prior research of it and trained a CNN with a multi-stage learning approach. The introduction of low cost depth camera and point cloud datasets led to new methods for grasp planning using RGN-D images to recover 3D geometry from point clouds. The subsequent CNNs used for grasp detection could be categorized into two classes roughly: two stage detector(TSD) and one stage detector(OSD). TSD contains a RPN(region proposal network) and a detector. The feature maps are extracted in the first stage and utilized by the detecting network in the second stage \cite{ROI_based}, \cite{hybrid_deep}, \cite{fully_CNN}. By contrast, OSD detects an object on each grid in a single stage \cite{real_time_CNN,yu2022se,cao2022efficient}.
\par
\textbf{Semantic segmentation }is a fundamental task in the area of computer vision(CV) which aims to assign each pixel of an image to its corresponding semantic class. \cite{garcia2017review,mo2022review,hao2020brief,ren2023visual} conclude a comprehensive context in this field. Owing to their property for labeling objects in images, this type of algorithms satisfy several needs in the real-world scenes: Feng et al. \cite{feng2020deep} examine the challenges involved in deep multi-modal object detection and semantic segmentation within autonomous driving scenarios; Flohr et al. \cite{flohr2013pedcut} present an iterative framework for achieving highly accurate pedestrian segmentation; Tao et al. \cite{tao2018automatic} propose a cascading network which transforms the input defect image into a pixel-wise prediction mask to detect metallic defects, being based on semantic segmentation. In this field, the traditional supervised algorithms greatly extend the cabibility of segmentation models. Therefore, such kind of methods such as random forest \cite{schroff2008object}, \cite{gao2016accurate} as well as visual grammar \cite{han2008bottom}. Nowadays, the rapid growth of deep learning leads to the advent of the era of CNNs. Long et al. \cite{long2015fully} proposed a fully convolutional network to achieve the outstanding outcomes in semantic segmentation, combining several contemporary classification CNNs(AlexNet \cite{AlexNet}, VGG \cite{Vgg} and GoogLeNet \cite{GoogLeNet}) and transfer learned representations of these networks by fine-tuning \cite{fine_tune}.

\section{\Large S\large{YSTEM}\Large{ O}\large{VERVIEW}}

The network's design, rooted in the methodologies presented in \cite{Por_segmen} for panoptic segmentation tasks, integrates thoughtful choices resulting in a lightweight and noise-resistant framework. One of the keystones of its streamlined design is the shared backbone, responsible for a singular extraction of features which it then feeds into the subsequent branches, efficiently eliminating redundant computations and optimizing memory use. This backbone, fortified by techniques like batch normalization, not only serves as the foundation for feature extraction but also plays a pivotal role in identifying graspable objects amidst clutter. Concurrently, the architecture's modules are optimized for peak performance while ensuring a lean footprint. Parameter sharing and regularization further enhance the compactness of the model.

\begin{figure*}[htbp]
   \centerline{\includegraphics[scale=0.4]{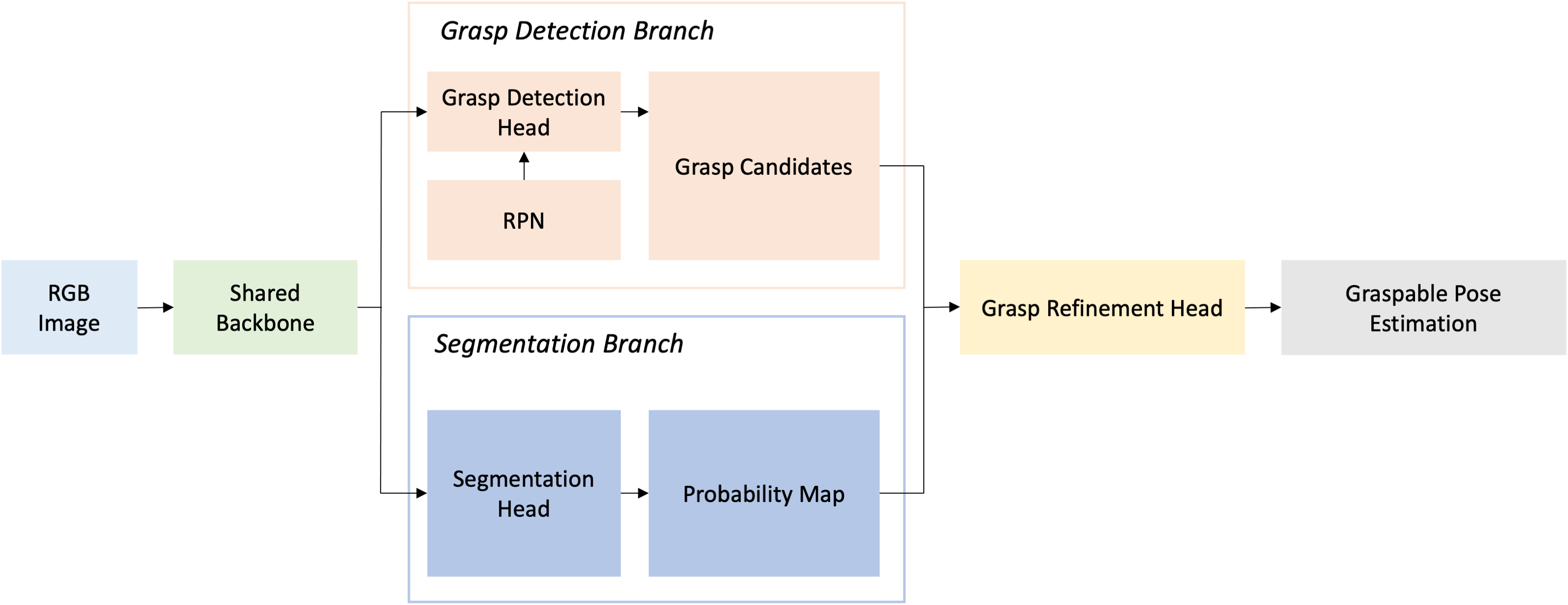}}
   \caption{Total architecture of the network. The grasp detection branch and the segmentation branch share the same backbone as the feature extractor, which learns variable information from the RGB images. Both outputs from two branches are used as input for the refinement head, which generates more precise grasp candidates.}
   \label{Fig2}
\end{figure*}

The semantic segmentation branch stands out as the vanguard against noise, diligently filtering out irrelevant background distractions and sharply delineating graspable objects. Building on this, the grasp refinement head fine-tunes the detection process. By assiduously combining and refining data, it sharpens the accuracy of the graspable object detection, ensuring that the robotic arm's approach towards an object is both precise and reliable. In tandem, the architecture's components work harmoniously to produce a system that's not only lightweight and agile but also displays an impressive resilience against noise. The culmination is a solution ideally suited for high-precision robotic applications in real-world, noise-rich environments. The overview of the whole architecture is shown in Figure \ref{Fig2}.
\subsection{\large{\textit{Shared Backbone}}}
We applied a slightly adapted ResNet-101 \cite{Res101} combining a Feature Pyramid Network(FPN) \cite{FPN} on the top of the whole network as the feature extractor. The modules {conv2, conv3, conv4, conv5} of ResNet are connected to the FPN. On top of that, all original Batch Normalization + ReLU layers of ResNet-101 are substituted for synchronized Inplace Activated Bath Normalization(InplaceABNSync) \cite{InplaceABNSync} with the LeakyReLU as activation function, optimizing memory utilization efficiency.

\subsection{\large{\textit{Grasp Detection Branch}}}
The employed architecture for grasp detection is based on the Faster R-CNN \cite{Fer-RNN}, which is the state-of-the-art object detector, incorporating a Region Proposal Network and a detection stage. The details of how we modified the object detector are as followed.
\par
\textit{1) Region Proposal Network}
The RPN accepts the features of the backbone network as input. The proposals are defined as $\hat{r}=(\hat{x},\hat{y},\hat{\omega},\hat{h})$ where $(\hat{x},\hat{y})$ represent center of the proposal in pixel coordinates and $(\hat{\omega},\hat{h})$  denote the width and height, respectively. Note that the region proposals are axis-aligned, lacking information regarding their potential orientation.\
\par
\textit{2) Grasp Detection Head}
The detection head predicts grasp candidates which are defined in Equation \ref{equ_g_define}. Each generated region proposal $\hat{r}$ is used as input for the detection head. After that, ROIAlign \cite{M-r-cnn} is used for extracting feature maps with 14 $\times$ 14 spatial resolution. Subsequently, average pooling with a kernel size of 2 is applied to each feature map, followed by their input into two fully connected (fc) layers, each consisting of 1024 neurons. Following each fc layer, an InPlaceABNSync normalization layer and Leaky ReLU activation with a slope of 0.01 are employed. The resulting outputs are then fed into two sub-networks:
\paragraph{Grasp orientation prediction.}
The first sub-network is composed of a fc layer with 1024 neurons followed by $N_{classes}+1$ output units, whereas $N_{classes}$ denotes the number of the orientation classes. Inspired by the work of \cite{Real_multi}, we discretize the grasp orientation $\theta$ into $N_{classes} = 18 $intervals with equal length, where each interval is represented by its mean value. The set containing total possible orientation classes is defined as $\mathcal{C}={(1,...,N_{classes})}$ including an additional class $\emptyset$ to evaluate the potential invalidity of a proposal. The ouput units generate logits for a softmax layer capturing the probability distribution across total potential orientation classes. The score function $s^{c}$ is calculated based on the probability assigned to orientation class $c \in C$.

\paragraph{Rectangular prediction}
The second sub-network consists of a fc layer with 1024 neurons which is the same as the first sub-network. However, the number of following output units is up to $4N_{classes}$. These output units encode class-specific correction factors $(t^{c}_{x},t^{c}_{y},t^{c}_{\omega},t^{c}_{h})$ for each input proposal $\hat{r}$. Afterwards, the defined grasp candidate $g$ can be directly calculated with the factors and the orientation information given by the score function $s^{c}$.
\subsection{\large{\textit{Segmentation Branch}}}
The segmentation head accepts the output of the first four scales generated by the FPN as its input. The FPN features of each level are individually input into a dedicated Mini-DeepLab module \cite{Por_segmen}. This architectural design enables the capturing of global structures in the input image while maintaining relatively low memory consumption. Subsequently, the output of each Mini-DeepLab module is up-sampled to $\frac{1}{4}$ of the input image size. All feature maps will be then concatenated and accepted by a 1 $\times$ 1 convolutional kernel with $S_{classes}$ output channels, which means the probability distribution of all semantic classes. Note that our proposed segmentation branch aims to provide prior information for the grasp pose refinement at the later stage, indicating that the result of segmentation branch includes the information of feasible grasp object candidates. To train our segmentation network, we manually relabel the dataset to provide additional feasibility information of the objects in the scenes. Hence, the segmentation branch is able to segment different classes of objects and only mask the feasible objects at the meantime, providing necessary features for the following refinement head.
\subsection{\large{\textit{Grasp Refinement Head}}}

\begin{figure*}[!htbp]
   \centerline{\includegraphics[scale=0.5]{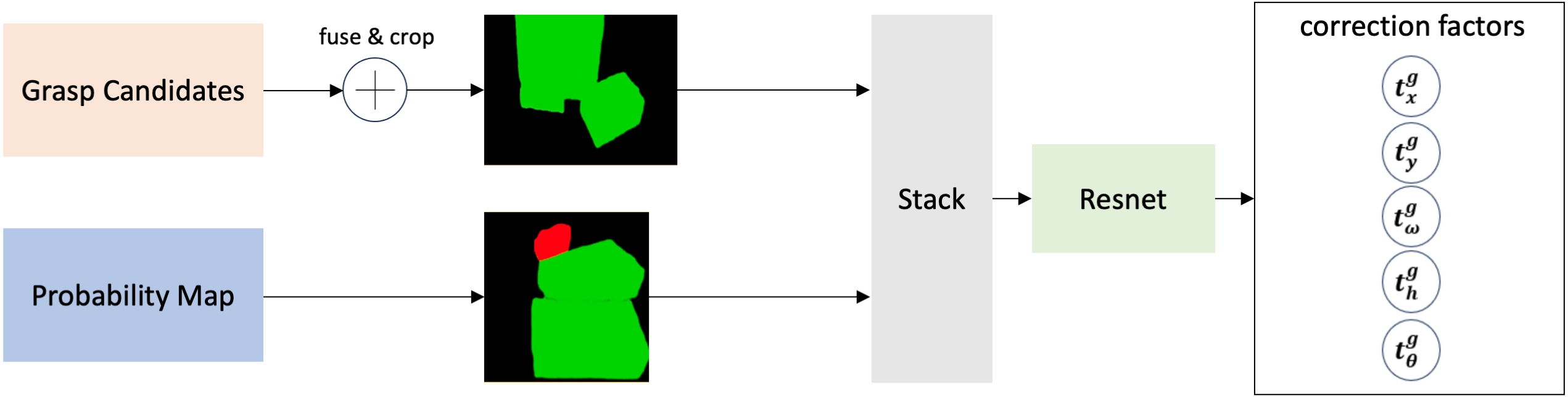}}
   \caption{The architecture of the grasp refinement head is based on ResNet-101 which takes as input two stacked semantic probability maps with dimensions (H $\times$ W), where (H, W) denotes the height and width of the probability map respectively. The output units of ResNet-101 represent refined correction factors $ (t_{x}^{g},t_{y}^{g}, t_{\omega}^{g}, t_{h}^{g}, t_{\theta}^{g} )$  which can be directly utilized to compute the refined grasp candidate. It is important to note that this operation is performed simultaneously for N grasp candidates, resulting in an input of dimension (N $\times $ 2 $\times $ H $\times $ W) for ResNet.}
   \label{Fig 3}
\end{figure*}

The grasp refinement head shown in Figure \ref{Fig 3} takes the previously calculated grasp candidates from the grasp branch and a semantic probability map generated by the semantic branch as input. To begin with, through cropping the area of the grasp candidate from the probability map, both information from different branches is fused. Later, the original probability map and the cropped ones are stacked together and accepted by the following ResNet-101 \cite{Res101}. This method allows us to merge geometric information of grasp candidates with information from segmentation branch about object shape. Following that, we employ an InPlaceABNSync normalization layer and Leaky ReLU activation function with a slope of 0.01. Note that each output denotes refined correction factors $(t_{x}^{g},t_{y}^{g},t_{\omega}^{g},t_{h}^{g},t_{\theta}^{g})$ for each grasp candidate $g$ to then calculate the refined candidate $\tilde{g}$.

\section{\Large{T}\large{RAINING }\Large{L}\large{OSSES}}
In order to simultaneously learn the tasks of grasp detection, semantic segmentation and grasp candidate refinement, we denote the merged loss function $L $ as
\vspace{-0.3cm}
\begin{equation}
   L=\lambda_{gr}L_{gr}+\lambda_{se}L_{se}+\lambda_{re}L_{re}
   \label{equ_loss_general},
   \vspace{-0.3cm}
\end{equation}

where $L_{gr}$ denotes the grasp detection loss, $L_{se} $ describes the semantic segmentation loss and $L_{re} $ represents the loss of grasp refinement candidates respectively. Each of them is assigned a weight, denoted by the hyperparameter $\lambda$.
\subsection{\large{\textit{Grasp Detection Loss}}}
The grasp detection loss $L_{gr}$ is defined as 
\begin{equation}
L_{gr}=L_{RPN}+L_{box}+L_{rot}
\end{equation}
where $L_{RPN}$ denotes the loss of the RPN, $L_{box}$ describes the regression loss for the bounding box prediction and $L_{rot}$ represents the classification loss for the grasp orientation $\theta$.
The loss function $L_{rot}$ is defined as
\begin{equation}
   L_{rot}=-\frac{1}{\lvert R \rvert}\sum_{\hat{r} \in R_{+}}logs_{\hat{r}}^{c_{r}}-\frac{1}{\lvert R \rvert}\sum_{\hat{r} \in R_{-}}logs_{\hat{r}}^{\emptyset},
\end{equation}   
where $R_{+}$ defines the set of valid region proposals generated  by the RPN, while $R_{-}$ defines the set of invalid region proposals respectively and $R$ denotes the union of them which means $R=R_{+}\cup R_{-}$. The score function $s_{\hat{r}}^{c_{r}}$ defines the likelihood that the region proposal $\hat{r}$ belongs to the ground truth orientation class $c_{r}$, while $s_{\hat{r}}^{\empty}$ represents the likelihood that the region proposal is considered invalid.
We employ the loss $L_{box}$ for box regression and it is defined as
\begin{equation}
   L_{box}=\sum_{i \in {x,y,h,\omega}}smooth_{L_{1}}(t_{i}^{c}-t_{i}^{*}),
\end{equation}
while the $smooth_{L_{1}}$ normalization function \cite{Fer-RNN} defined as
\[
\text{smooth}_{L_{1}}(x)=
\begin{cases}
0.5x^{2}, & \text{if } |x| < 1 \\
|x|-0.5, & \text{otherwise}
\end{cases}
\]
the parameter $t_{i}^{c}$ \cite{rich_fea} denotes the results of the bounding box prediction. The other parameter $t^{*}_{i}$ is defined as 
\[
\begin{cases}
t_{x}^{*}=\frac{x^{*}-\hat{x}}{\hat{w}}, \\
t_{y}^{*}=\frac{y^{*}-\hat{y}}{\hat{w}}, \\
t_{\omega}^{*}=log(\frac{\omega^{*}}{\hat{\omega}}), \\
t_{h}^{*}=log(\frac{h^{*}}{\hat{h}}), \\
\end{cases}
\]
whereas $(\hat{x},\hat{y},\hat{\omega},\hat{h})$ and $(x_{*},y_{*},\omega_{*},h_{*})$ denote parameters from the region proposal $\hat{r}$ and ground truth candidates respectively.
\subsection{\large{\textit{Segmentation Loss}}}
We also denote $\mathcal{S}=\{1,...{S_{classes}}\}$ as the set of semantic segmentation classes. Inspired by the work in \cite{Por_segmen} , the weighted loss is defined as 
\begin{equation}
   L_{se}=-\sum_{j,k}\omega_{j,k}logP_{j,k}(S_{j,k}),
\end{equation}
where $(j,k)$ means the pixel position in the image, $S_{j,k} \in \mathcal{S}$ denotes the semantic segmentation ground truth and $P_{j,k}(s)$ defines the predicted likelihood for the same pixel to be confirmed as a specific semantic class $s \in \mathcal{S}$. The weights $\omega_{j,k}$ are generated by the semantic segmentation branch to implement a pixel-wise hard negative mining, which chooses the lowest predicted likelihood $P_{j,k}$ for all $(j,k)$ pixel using $\omega_{j,k}=\frac{4}{WH}$, and $\omega_{j,k}=0$ otherwise. Note that $W,H$ means the width and height of the given image respectively.

\section{\Large{E}\large{XPERIMENTS AND }\Large{R}\large{ESULTS} }
We evaluates the advantages of our deployed method on OCID dataset \cite{EasyLabel}. We also compare our proposed structure with the state-of-the-art methods. We initialize the feature extractor network with pretrained weights of ResNet101 \cite{Res101} and freeze the parameters of first two network modules {conv1, conv2} within the whole training turns. Note that we used a Nvdia GTX 3080 graphics card to run all training and evaluation turns.
\subsection{\large{\textit{Evaluation Metric}}}
To compare with the results from the other methods, we deploy the popular Jaccard index as accuracy measurement. A grasp candidate is regarded as correct when:
\begin{enumerate}
   \item The discrepancy in rotation angles between the output grasp detection rectangle ($g_{p} $) and the ground truth rectangle $ g_{gt}$ is less than $30\degree $;
   \item The Intersection over Union(IoU) of them is greater than 0.25 which means:
   \begin{equation}
      \mathit{IoU}=\frac{\lvert g_{p} \cap g_{gt}\rvert}{\lvert g_{p} \cup g_{gt} \rvert},
      \label{IoU_define}
   \end{equation}  
   For semantic segmentation, we report the IoU between predicted ground truth segmentation.
\end{enumerate}
\subsection{\large{\textit{Data Preprocessing}}}
\paragraph{OCID Dataset with Grasp-Extension}
In the real sceneries, the images for the robots are always blurry due to the distortion and unstability of the embedded cameras in the robots. To simulate the scenes in real world as closely as possible, we add some Gaussian noise or salt-and-pepper noise to the images, which also evaluate the robustness of the network.

The training phase spanned 800 epochs with a batch size of 10 for each training iteration. Impressively, our model achieved a grasp accuracy of 89.02\% and an Intersection over Union (IoU) score of 94.05\%, highlighting the efficacy of our architecture in grasping tasks and the precision of object overlap and detection, respectively.

A standout feature of our system is its processing capability, clocking in at an impressive 15.2 images per second. This expedited processing speed can be attributed to several factors intrinsic to our design. The efficient distribution of parameters across the network's modules ensures that computational resources are not overburdened. Additionally, the lightweight nature of our architecture, stemming from strategic choices like a shared backbone and streamlined modules, minimizes computational overhead. The system's agility in handling and processing information, combined with optimizations tailored for speed, allows for such brisk image processing rates. This makes our solution not only theoretically robust but also practically adept for real-time applications, particularly in dynamic robotic environments where swift decision-making is paramount.

\begin{figure*}[htp]
	\centering
	\begin{subfigure}[b]{0.3\linewidth}
		\includegraphics[width=\linewidth, height=3cm]{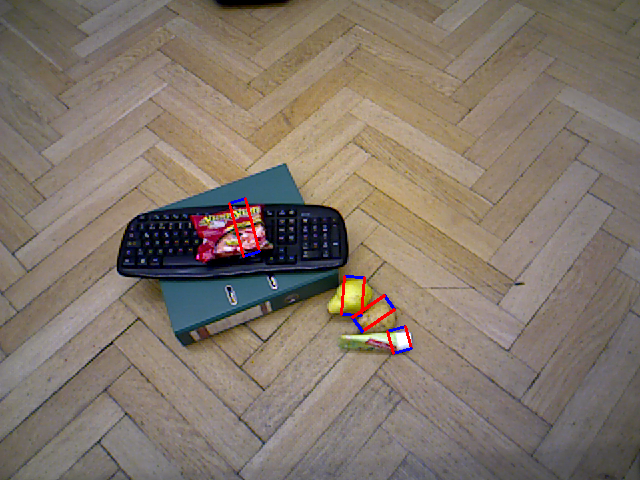}
	\end{subfigure}
	\begin{subfigure}[b]{0.3\linewidth}
		\includegraphics[width=\linewidth, height=3cm]{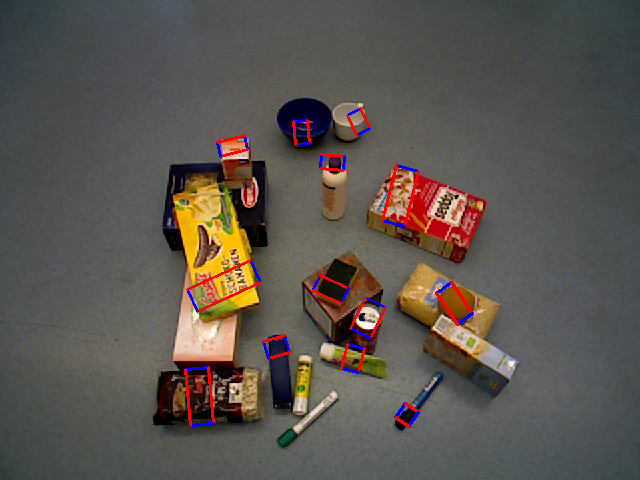}
	\end{subfigure}
	\begin{subfigure}[b]{0.3\linewidth}
		\includegraphics[width=\linewidth, height=3cm]{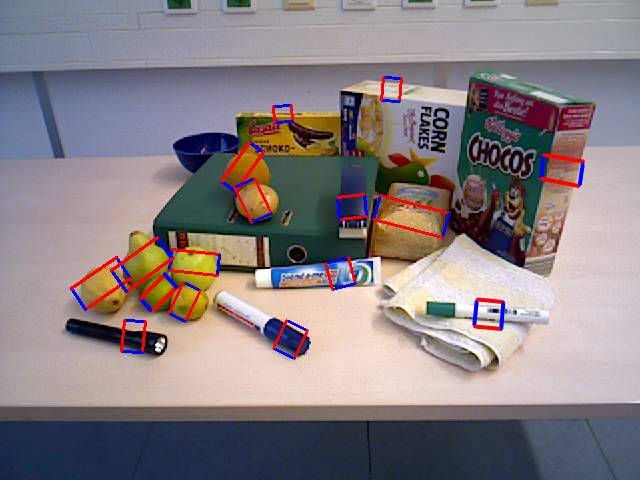}
	\end{subfigure}\\
	\begin{subfigure}[b]{0.3\linewidth}
		\includegraphics[width=\linewidth, height=3cm]{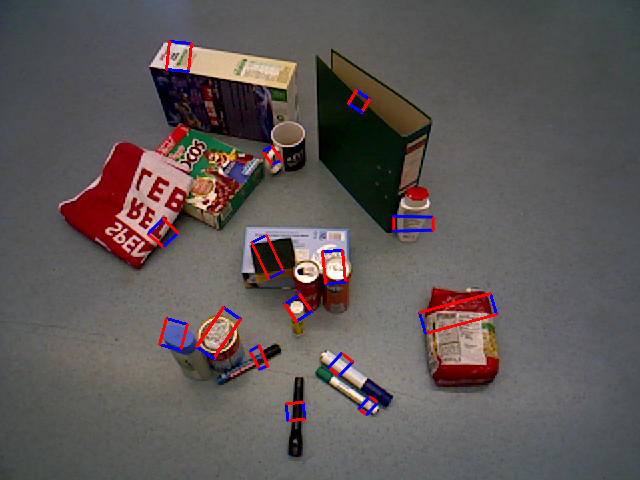}
	\end{subfigure}
	\begin{subfigure}[b]{0.3\linewidth}
		\includegraphics[width=\linewidth, height=3cm]{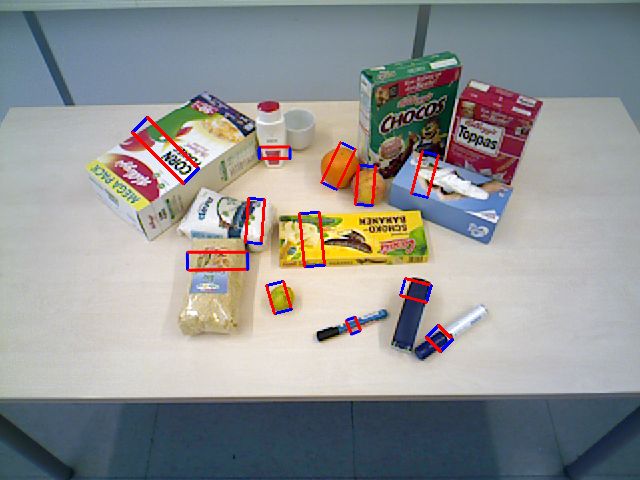}
	\end{subfigure}
	\begin{subfigure}[b]{0.3\linewidth}
		\includegraphics[width=\linewidth, height=3cm]{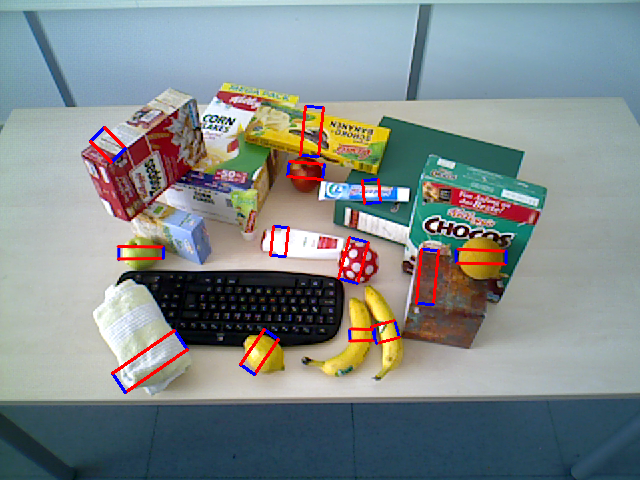}
	\end{subfigure}\\
	\caption{Results of grasp candidates in various scenes.}
	\label{Fig_suc}
\end{figure*}

In our system, the segmentation branch has been judiciously adapted to assess the feasibility of objects, specifically for the refinement head. Consequently, the final outputs predominantly display those grasp candidates deemed viable. In contrast, candidates associated with occluded objects or those supporting other entities are systematically filtered out from the concluding results. Figure \ref{Fig_suc} shows the results of the proposed model.
\begin{figure*}[ht]
	\centering
	\begin{subfigure}[c]{0.3\linewidth}
		\includegraphics[width=\linewidth, height=3.4cm]{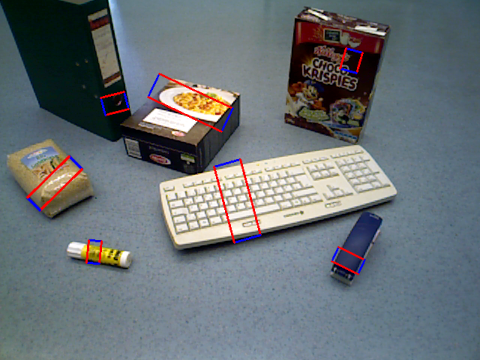}
	\end{subfigure}
	\begin{subfigure}[c]{0.3\linewidth}
		\includegraphics[width=\linewidth, height=3.4cm]{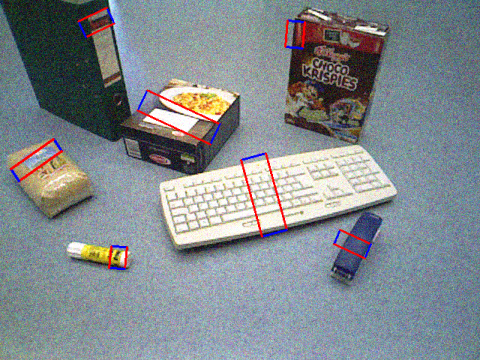}
	\end{subfigure}
	\begin{subfigure}[c]{0.3\linewidth}
		\includegraphics[width=\linewidth, height=3.4cm]{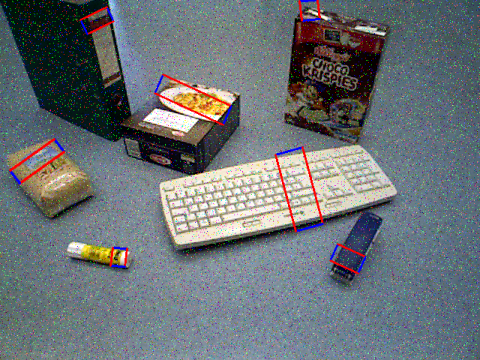}
	\end{subfigure}\\
	\begin{subfigure}[c]{0.3\linewidth}
		\includegraphics[width=\linewidth, height=3.4cm]{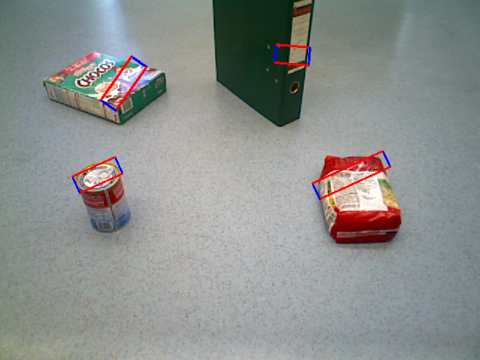}
	\end{subfigure}
	\begin{subfigure}[c]{0.3\linewidth}
		\includegraphics[width=\linewidth, height=3.4cm]{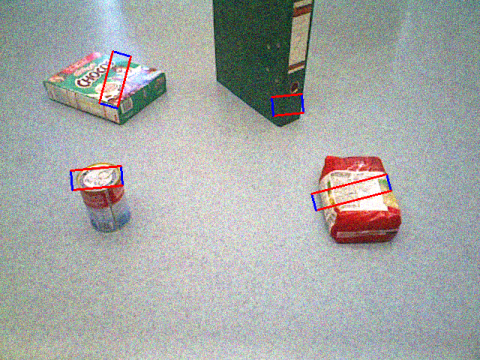}
	\end{subfigure}
	\begin{subfigure}[c]{0.3\linewidth}
		\includegraphics[width=\linewidth, height=3.4cm]{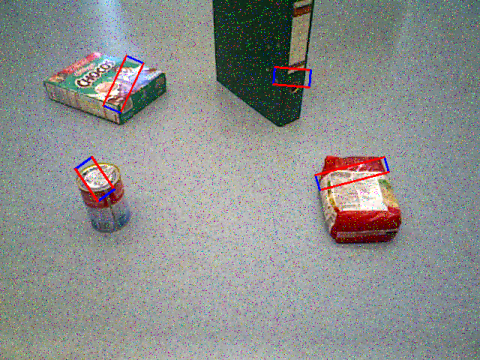}
	\end{subfigure}\\
	\begin{subfigure}[c]{0.3\linewidth}
		\includegraphics[width=\linewidth, height=3.4cm]{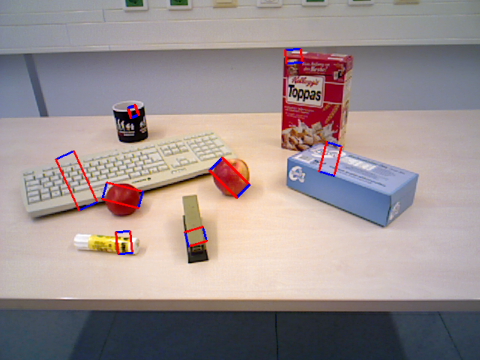}
	\end{subfigure}
	\begin{subfigure}[c]{0.3\linewidth}
		\includegraphics[width=\linewidth, height=3.4cm]{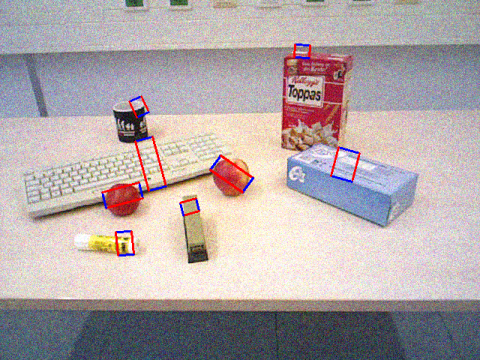}
	\end{subfigure}
	\begin{subfigure}[c]{0.3\linewidth}
		\includegraphics[width=\linewidth, height=3.4cm]{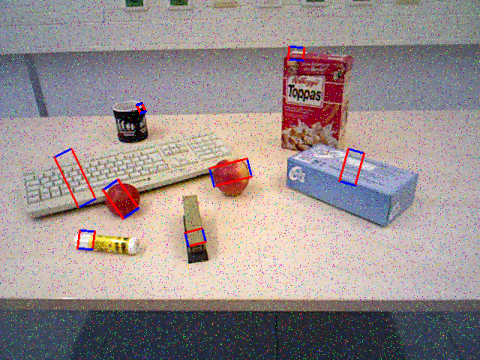}
	\end{subfigure}\\
	\begin{subfigure}[c]{0.3\linewidth}
		\includegraphics[width=\linewidth, height=3.4cm]{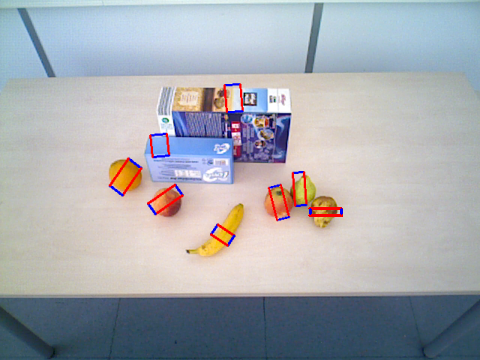}
	\end{subfigure}
	\begin{subfigure}[c]{0.3\linewidth}
		\includegraphics[width=\linewidth, height=3.4cm]{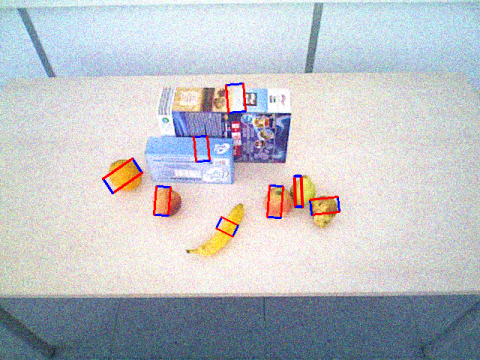}
	\end{subfigure}
	\begin{subfigure}[c]{0.3\linewidth}
		\includegraphics[width=\linewidth, height=3.4cm]{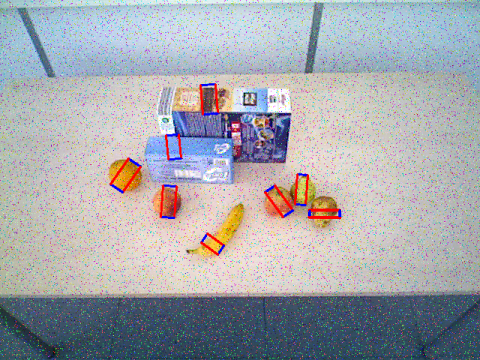}
	\end{subfigure}\\
	\caption{Results of grasp candidates. The input from left to right column: original images, images with Gaussian noise, and images with salt-and-pepper noise.}
	\label{Fig4}
\end{figure*}
\subsection{\large{\textit{Results}}}
As mentioned above, to enhance the stability of the network when facing the noise in the input image, we add some noise to the dataset and train the network with the noisy images. Figure \ref{Fig4} has shown results with different types of noise(non-noise, Gaussian noise and salt-and-pepper noise). It is revealed by the figure that when facing three kinds of noise, the results seem slightly discrepant, especially about the unfeasible grasp detection. For some results with noise, some obscured objects are also detected and some objects are even with more than one grasp candidates which is worth studying.

\subsection{\large{\textit{Limitations}}}
\begin{figure*}[htbp]
	\centering
	\begin{subfigure}[b]{0.3\linewidth}
		\includegraphics[width=\linewidth]{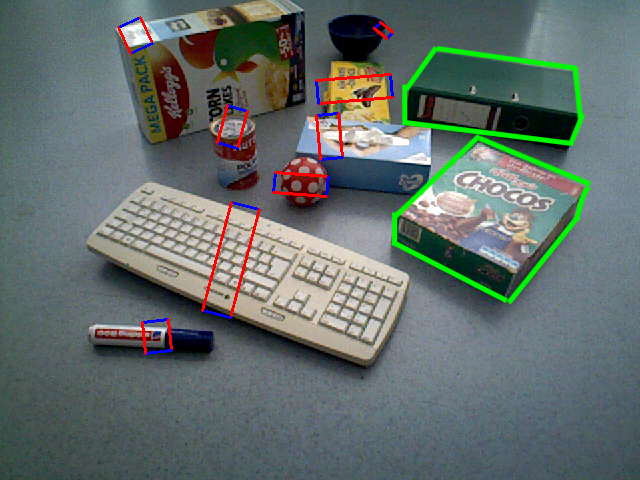}
	\end{subfigure}
	\begin{subfigure}[b]{0.3\linewidth}
		\includegraphics[width=\linewidth]{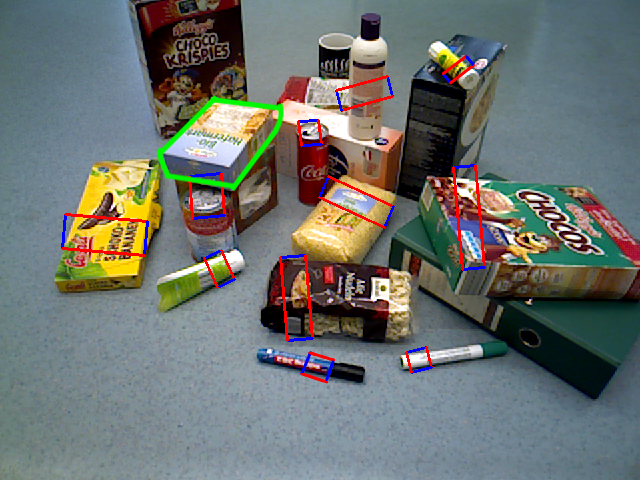}
	\end{subfigure}
	\caption{The failure cases of the deployed network, which are highlighted in green lines.}
	\label{Fig_failure}
\end{figure*}

While our network demonstrates notable capabilities, it does present certain limitations. As depicted in Figure \ref{Fig_failure}, the system occasionally generates unfeasible grasp poses or multiple grasp poses for a singular object. A consistent challenge arises with larger objects, such as boxes, which are frequently overlooked by the network. Often acting as foundational supports for smaller items, these larger entities are typically labeled as ungraspable in the ground truth. Consequently, the network tends to misjudge them as ungraspable even in instances when they aren't overlaid by smaller objects.
\section{\Large{C}\large{ONCLUSIONS}}
In conclusion, the field of robotics has witnessed significant strides in automating object grasping, with mechanical-arm-equipped robots being at the forefront of these advancements. As we transition from traditional depth sensor-dependent methods to leveraging the ubiquity of standard RGB images, we embrace both the potential and challenges inherent in this shift. RGB images, though cost-effective, bring forth a myriad of complications, not least due to their lack of depth information and vulnerability to various noise sources. 

To summarize, the key takeaways from our research are threefold: the development of a trainable neural network adept at detecting graspable objects from RGB images, the inception of a modular design integrating multiple training facets for enhanced grasp detection, and the crafting of a robust system proficient at deciphering blurry and noise-afflicted visuals. As the robotics domain continues its rapid evolution, it's paramount that the solutions we devise are both innovative and adaptive, and we believe our network epitomizes this ideal.

Our work has opened up interesting directions for future research. First, while our current architecture excels at grasping based on RGB imagery, integrating other sensory inputs, such as depth or tactile feedback, could potentially further enhance the system's robustness and versatility in diverse environments. Second, our results indicated commendable performance against specific types of noise. Investigating advanced noise-filtering algorithms, especially for unexpected or rare noise patterns, would ensure the system's resilience in even more challenging environments. Third, while our current distribution of parameters has yielded efficient results, further refinements in the allocation of computational resources can be explored. This would involve a deeper dive into each module's individual contribution to the overall processing time and accuracy. Lastly, implementing an adaptive learning mechanism where the network adjusts its parameters based on real-time feedback could make the system more autonomous and responsive to dynamic environments.

\bibliographystyle{IEEEtran}
\bibliography{main.bib}

\end{document}